\documentclass[letterpaper, 10 pt, conference]{IEEEtran}
\IEEEoverridecommandlockouts

\usepackage{cite}
\usepackage{amsmath,amssymb,amsfonts}
\usepackage{algorithmic}
\usepackage{graphicx}
\usepackage{textcomp}
\usepackage{xcolor}
\usepackage{multirow}  
\usepackage{stfloats}
\usepackage{geometry}
\title{\LARGE \bf
Jacquard V2: Refining Datasets using the Human In the Loop Data Correction Method
}

\author{Qiuhao Li$^{1*}$ and Shenghai Yuan$^{2}$
\thanks{* Corresponding Author.}
\thanks{$^{1}$Qiuhao Li is with College of Robotics Science and Engineering, Northeastern University, Shenyang City, Liaoning Province, China 110167, 
        {\tt\small li.qiuhao@outlook.com, liqiuhao@stumail.neu.edu.cn}}%
\thanks{$^{2}$Shenghai Yuan is with School of Electrical and Electronic Engineering, Nanyang Technological University, 50 Nanyang Avenue, Singapore 639798, 
        {\tt\small shyuan@ntu.edu.sg}}%
}

\begin{document}
\maketitle
\thispagestyle{empty}
\pagestyle{empty}

\begin{abstract}
In the context of rapid advancements in industrial automation, vision-based robotic grasping plays an increasingly crucial role. In order to enhance visual recognition accuracy, the utilization of large-scale datasets is imperative for training models to acquire implicit knowledge related to the handling of various objects. Creating datasets from scratch is a time and labor-intensive process. Moreover, existing datasets often contain errors due to automated annotations aimed at expediency, making the improvement of these datasets a substantial research challenge. Consequently, several issues have been identified in the annotation of grasp bounding boxes within the popular Jacquard Grasp Dataset\cite{depierre2018jacquard}. 
We propose utilizing a Human-In-The-Loop(HIL) method to enhance dataset quality. This approach relies on backbone deep learning networks to predict object positions and orientations for robotic grasping.
Predictions with Intersection over Union (IOU) values below 0.2 undergo an assessment by human operators. After their evaluation, the data is categorized into False Negatives(FN) and True Negatives(TN). FN are then subcategorized into either missing annotations or catastrophic labeling errors.
Images lacking labels are augmented with valid grasp bounding box information, whereas images afflicted by catastrophic labeling errors are completely removed. The open-source tool Labelbee was employed for 53,026 iterations of HIL dataset enhancement, leading to the removal of 2,884 images and the incorporation of ground truth information for 30,292 images.
The enhanced dataset, named the Jacquard V2 Grasping Dataset, served as the training data for a range of neural networks. We have empirically demonstrated that these dataset improvements significantly enhance the training and prediction performance of the same network, resulting in an increase of 7.1\% across most popular detection architectures for ten iterations. This refined dataset will be accessible on Google Drive and Baidu Netdisk, while the associated tools, source code, and benchmarks will be made available on GitHub (https://github.com/lqh12345/Jacquard\_V2).

\end{abstract}


\section{Introduction}

In robot visual grasping, object recognition by computer vision is pivotal. A high-quality dataset is vital for training effective predictive models, directly influencing their accuracy. Deep learning networks rely on ground truth grasp box coordinates in datasets to optimize the model weights. As the entire robotic arm's trajectory hinges on these coordinates, even a tiny pixel error in ground truth can determine success or failure. Precision in ground truth annotations is paramount, as only accuracy ensures reliable model training.

\begin{figure}[t]
 \centering         
 \includegraphics[width=\linewidth]{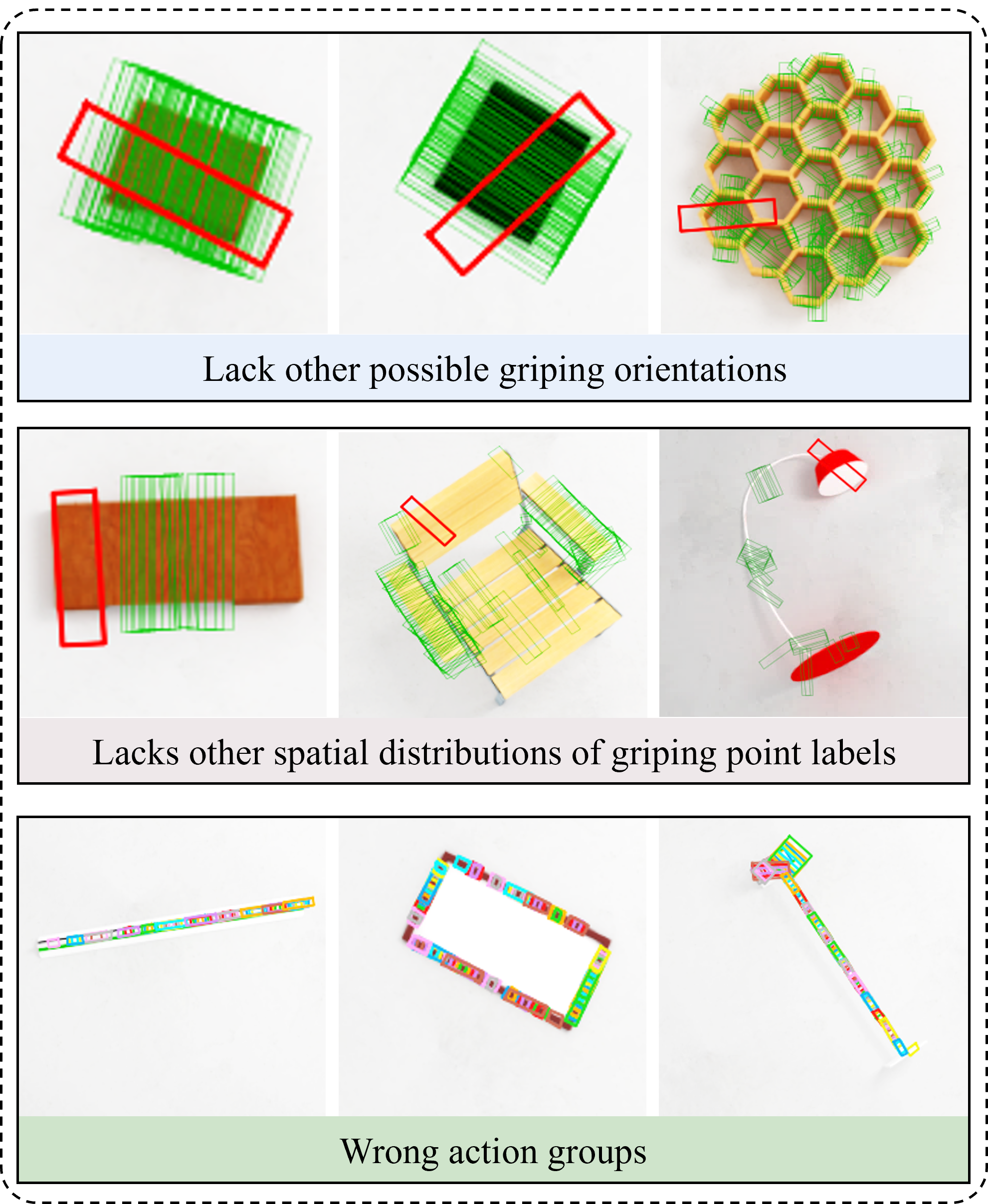}
 \vspace{-1em}
 \caption{Challenges arise within the Jacquard Grasp Dataset\cite{depierre2018jacquard}. The green markers represent the dataset labels, while the red markers indicate predictions. Although the predictions are frequently accurate, they may sometimes be incorrectly labeled as incorrect due to discrepancies with the incomplete annotations.}
 \label{fig.p1}
\end{figure}

Creating new grasping datasets is typically demanding and costly. While computer-aided methods offer speed, they are prone to errors. Manual annotation, on the other hand, reduces errors but demands more time and resources. In both approaches, some degree of error is unavoidable. Therefore, improving the accuracy of ground truth annotations in existing datasets is crucial, particularly when the current dataset exhibits biases in object handling, such as the case with sticks, as depicted in Figure. \ref{fig.p1}.

To address these challenges, we present a rapid HIL approach to enhance the Jacquard V2 dataset swiftly and effectively. Without altering the network architecture, we focus solely on improving annotations. This refinement results in a notable 5-8 \% average increase in performance. To validate the efficacy of our proposed solution, we evaluate it using various open-source architectures. The contributions of this paper can be summarized as follows:

\begin{itemize}
\item We propose an innovative HIL method to improve the Jacquard Grasp Dataset\cite{depierre2018jacquard}. By quickly selecting pseudo-label annotations generated from backbone models, this approach corrects errors, enhances data consistency, and ensures accurate grasping box information, all without manual drawing actions.

\item  We thoroughly benchmark all popular methods to prove that better annotation can consistently improve the performance of various networks throughout all improvement iterations. Our research also unveils the influence of network complexity on learning outcomes.

\item Our approach involves error correction, incorporates human intelligence, and streamlines dataset refinement for cost and time efficiency. We will release the improved dataset as open-source, along with benchmark codes, to encourage collaboration and advancements in the field.
\end{itemize}

\subsection{Dataset Issues Characterization}
The Jacquard Grasp Dataset\cite{depierre2018jacquard} is generated using Automatic CAD verification methods. In this study, we visually examined the annotated results of the Jacquard Grasp Dataset and identified three significant issues:

\textbf{Lack of other possible griping orientations}: One prominent issue is that a subset of images exhibits a single-directional grasp box orientation. For instance, some images only depict horizontal grasping, while others exclusively show vertical grasping. This discrepancy introduces erroneous information into the learning process of the network, hindering the training of generalizable intelligent agents. The underlying cause of this problem lies in the failure of the computer network responsible for generating ground truth annotations to effectively capture the diversity of graspable objects.

\textbf{Lacks other spatial distributions of griping point labels}: Another issue manifests in certain images where the grasp box coverage is incomplete. Due to the inherent uncertainty in the computer-generated positioning of grasp boxes, large areas in some images that should have been covered by grasp boxes remain unannotated. Such incomplete coverage renders the dataset incomplete and inadequate for training purposes. 

\begin{figure}[t]
    \centering         
    \includegraphics[width=\linewidth]{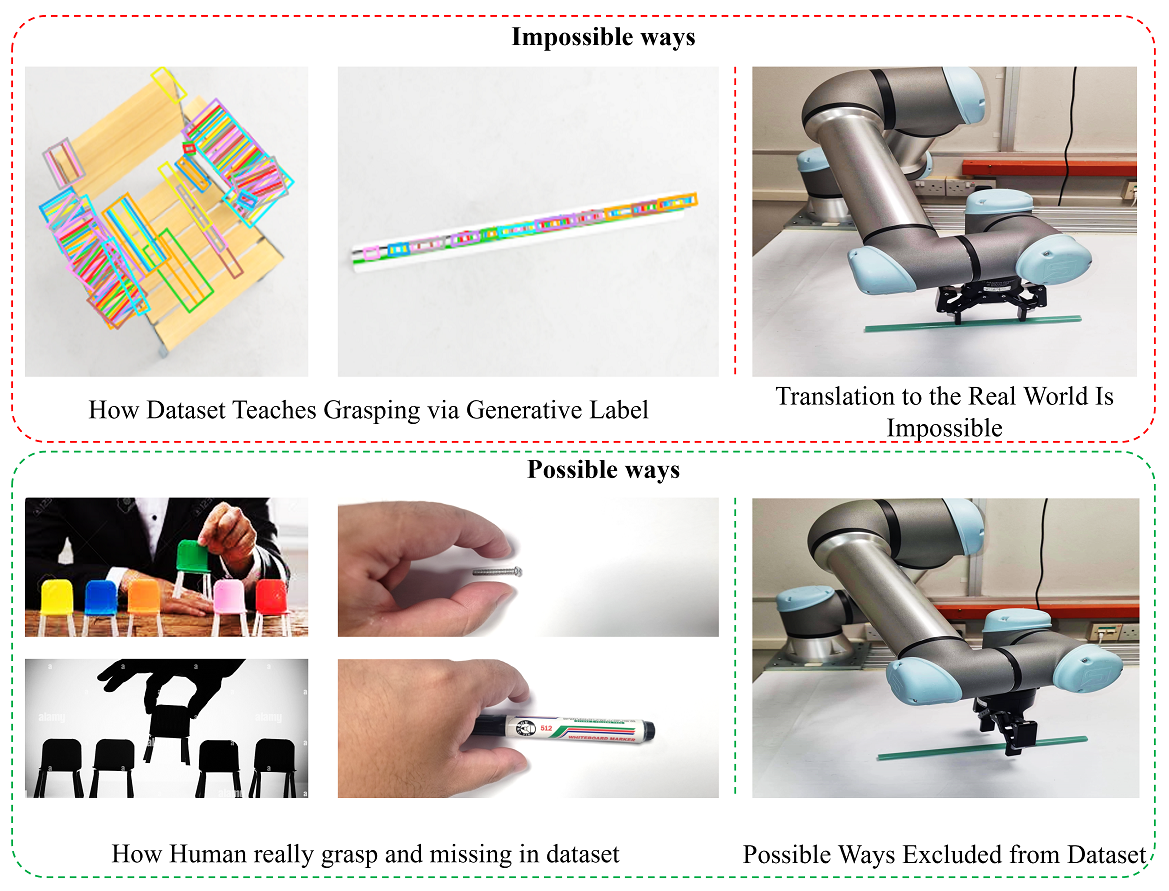}
    \caption{Top left corner: Dataset annotations,
Top right: Visualization of how this dataset translates to real robot manipulation,
Bottom left: Human interaction with the same classes,
Bottom right: Verification that the human native approach is correct.}
    \label{fig.p2}
\end{figure}

\textbf{Wrong action groups}: In Fig. \ref{fig.p2}, some images in the dataset contain incorrect grasp boxes. For instance, objects such as pens or chopsticks have parallel grasp boxes, which contradicts the logical placement for grasping. Consequently, the end position of the robotic arm needs to penetrate these objects to lift them, which is inconsistent with practical scenarios. These errors create a mismatch between the neural network and the dataset, requiring the network to adapt to the annotation preferences specific to the Jacquard Grasp Dataset to achieve high scores. To address these issues, it is crucial to remove datasets with a significant number of erroneous labels.

The manual annotation of datasets yields high precision but suffers from slow implementation, making it unsuitable for large-scale dataset annotation.

\section{Related Work}


In the field of robotic visual grasping, several publicly available datasets have been developed, including the Dex-Net 2.0 Dataset \cite{mahler2017dex}, YCB-Video Dataset\cite{xiang2017posecnn}, Willow Garage Object Dataset\cite{rusu2010semantic}, Willow Object Dataset\cite{tang2012textured}, Big Bird Dataset\cite{zaheer2020big}, TUM Kitchen Dataset\cite{tenorth2009tum}, JHU-ISI Gesture and Skill Assessment Working Set\cite{gao2014jhu} and Jacquard Grasp Dataset\cite{depierre2018jacquard}. These datasets have played a crucial role in training neural networks for robotic grasping tasks. However, to maximize the effectiveness of these datasets, various data augmentation techniques have been proposed.

Although commonly used data augmentation methods  can partially expand the dataset and mitigate the impact of dataset limitations, they are inadequate in addressing issues such as erroneous labels and missing annotations. These problems persist even after the implementation of data augmentation methods and continue to affect the accuracy of the dataset.

Numerous automated approaches have been proposed for generating annotations \cite{castrejon2017annotating,nguyen2022ntu}, primarily tailored to well-defined problems like semantic segmentation and tracking. In contrast, the challenge of object grasping is inherently ambiguous, with an infinite range of ways to manipulate objects, making it a complex problem to define precisely. While various weakly supervised methods \cite{bruls2018mark} have also been suggested, these solutions are not readily applicable to the complexity of grasping.

One potential approach to tackle this ambiguity is through label fusion techniques \cite{marion2018label}, although these methods require prior knowledge of the objects involved. Another avenue explored involves training feasible paths applicable to grasping \cite{barnes2017find}, but this approach may falter when dealing with datasets containing thousands of unknown scenes.

Some researches\cite{bousmalis2018using,eppner2021acronym} have advocated for the utilization of simulations to generate potential dataset annotations. Nonetheless, it's worth noting that errors in the dataset may have originated from the simulation itself, raising questions about its reliability. Alternatively, some have suggested leveraging crowd-sourced knowledge to train neural networks for annotation tasks \cite{kappler2015leveraging}. However, placing full trust in the quality of crowd-sourced information can be a precarious endeavor.

Other works\cite{suchi2019easylabel} have introduced a technique that links camera motion with RGBD images to label objects automatically. Nevertheless, this method may not readily suit pre-existing static image datasets. Furthermore, various generative models have been proposed as potential solutions to the annotation challenge \cite{handa2016scenenet}. However, it is important to note that these generative models commonly face the well-known issue of hallucination. Addressing this concern is imperative to ensure the models can produce valid annotations rather than generating more erroneous labels.

\begin{figure*}[htbp]
 \centering         
\includegraphics[width=1.0\linewidth]{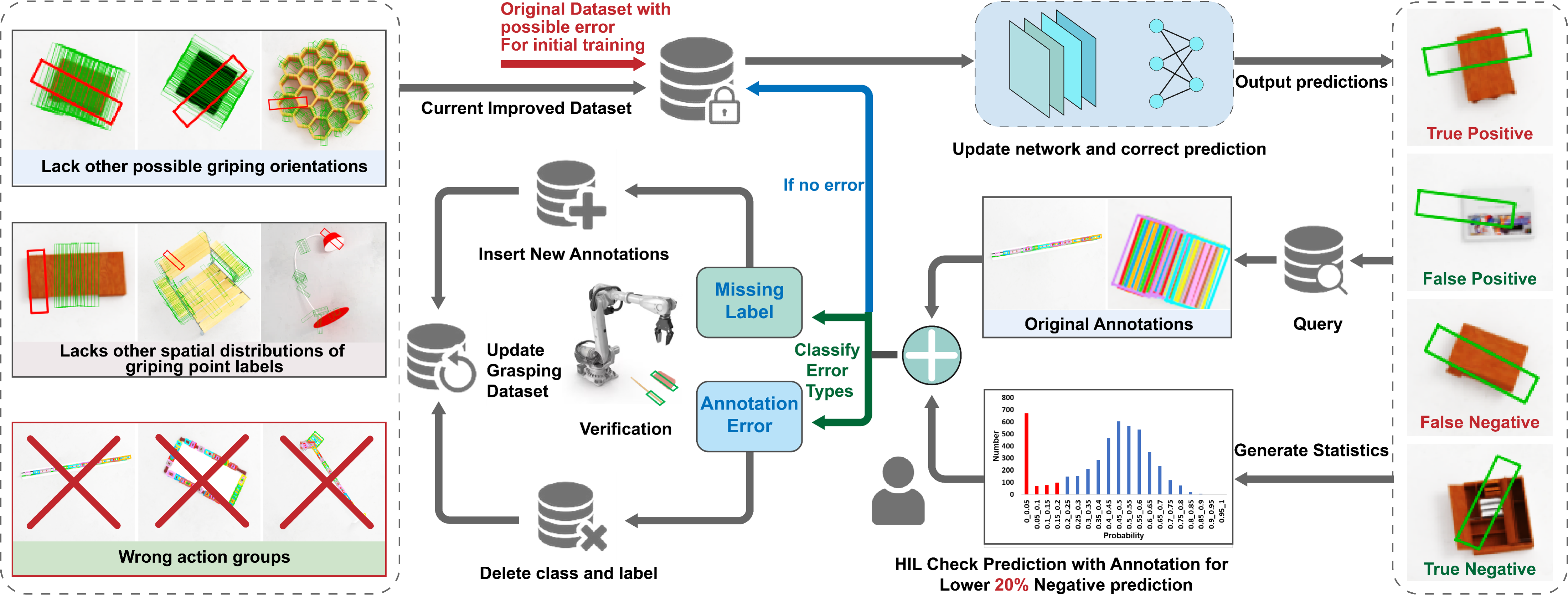}
 \vspace{-1em}
 \caption{Overall Workflow of the Dataset Refinement Method based on HIL.}
 \label{fig.p3}
 \vspace{-2em}
\end{figure*}

In many real-world engineering applications, achieving objectives with limited datasets is a common challenge. In such cases, deep neural networks often suffer from insufficient training data, leading to overfitting on the training set and poor generalization on the test set. Over the years, various techniques have been developed to mitigate the issue of overfitting, including geometric transformations, color transformations, masking, dropout \cite{krizhevsky2012imagenet}, batch normalization \cite{ioffe2015batch}, batch renormalization \cite{ioffe2017batch}, and layer normalization \cite{ba2016layer}. While these methods can reduce the occurrence of overfitting and mitigate the impact of imperfect datasets to some extent, they are unable to address the problems of erroneous labels and missing annotations. These issues persist even after the implementation of the aforementioned techniques and continue to affect the accuracy of the dataset.

Tian Tan et al.\cite{tan2023human} introduced a novel dataset refinement method for robotic grasping. It involves human-robot interaction to improve grasp efficiency by obtaining numerical grasp quality input from humans. Deep learning and computer vision are used to detect objects, and human operators rate grasp points. However, this method, while addressing erroneous labels, is intricate and resource-intensive, with varying criteria for grasp point scoring, potentially affecting its practical effectiveness. Additionally, it is susceptible to subjective human bias, as each person may have preferences for different grasping points.

In comparison to the aforementioned approaches, our proposed method addresses the issues of erroneous labels and missing annotations in the dataset. Moreover, human operators only need to judge the grasping ability of the objects in each image and assess the accuracy of the ground truth annotations for the grasp bounding boxes. This data refinement process is simple, yields stable results, and is easier to learn from.

\section{Proposed Method}

\subsection{Overall Overview of the Dataset Refinement Method based on HIL}
Despite the growing availability of robotic grasping datasets, researchers have encountered challenges in achieving high accuracy due to variations in the regions of interest within the main network and erroneous ground truth annotations. When experimental results are affected by the imperfections in the dataset, it hampers the accurate evaluation of the performance of new networks. To address this issue, we propose a HIL dataset refinement method.

We propose a HIL dataset refinement method that directly operates on the dataset, aiming to significantly enhance the recognition performance of models. The proposed system initially trains a model using the original dataset and generates grasp bounding boxes for each image. Subsequently, images with generated grasp bounding boxes having an Intersection over Union (IOU) less than 0.2 with the ground truth are selected and labeled as "False". Human operators then categorize all False images into two categories: False Negatives(FN) and True Negatives(TN), based on simulation and their own expertise. If an image is classified as an FN, human operators further categorize it into "Missing Label" or "Annotation Error". For images classified as Missing labels, valid grasp bounding box information is supplemented into the dataset. For images categorized as Annotation Errors, they are entirely removed from the dataset. Once all False images have been traversed, the dataset updating process is considered complete. A new model is then trained using the updated dataset, and the aforementioned steps are repeated, enabling cyclic improvement and updates to the dataset. Our proposed method offers a simpler and more efficient operational workflow, directly impacting the dataset and significantly improving the recognition effectiveness of the models.

\subsection{Training Losses Formulation}
Regarding the loss formulation, our approach largely aligns with the guidelines established by GG-CNN\cite{morrison2018closing}, with some slight adjustments. Instead of predicting multiple parameters such as grasping center, grasp angle, grasp width, and grasp quality, we focus on predicting only the center, angle, and width. We define a grasp point $g$ as 

\begin{center}
 $g = (o, \phi,\omega)$
\end{center}

where $o$ represents the grasping point's center, $\phi$ defines the orientation of the manipulator end effector, and $\omega$ denotes the width of gripper.
To ensure a smoother and more manageable distribution for the network to learn from, we decompose the angle into both a cosine term and sine terms, both in the range of $[-1,1]$. This decomposition\cite{hara2017designing} effectively eliminates any potential discontinuities in the data that would otherwise occur when the angle wraps around ±$\frac{\pi}{2}$ if the raw angle was directly used. In this case, the grasp point that the network learned can be defined as: 

\begin{center}
 $g = (o, c_{\phi}, s_{\phi},\omega)$
\end{center}

The losses are defined as the sum of squared differences between predicted and target values. This entails calculating the squared difference for each corresponding element, summing the entire tensor, and dividing by the number of elements to obtain the mean squared error. 
\begin{align*}
 {L}_{\text{overall}} = {L}_{o} + {L}_{c_{\phi}} + {L}_{s_{\phi}} + {L}_{\omega}
\end{align*}

The angle is recalculated from the sine and cosine terms to generate the grasping points.

\section{Experiment}

This section describes the experimental setup and experimental procedure used in this study.

\subsection{Experimental Design}

All experiments in this paper were conducted under the following environment:

Operating System: Ubuntu 16.04.7 LTS

CPU: Intel(R) Xeon(R) Silver 4210R CPU @ 2.40GHz

GPU: 1 x 3090TI on Cuda 11.1

Python Version: 3.7.12

Torch Version: 1.10.0

Torchvision Version: 0.11.0

For the HIL iteration process, we utilized the MobileNet V2\cite{sandler2018mobilenetv2} as the backbone network to generate the pseudo-labels. Each training session consisted of 100 epochs, with a batch size 32. The initial learning rate was set to 0.0005, and a dropout rate of 0.05 was employed.

For this dataset, we performed ten iterations of the HIL revisions process. After several revisions, images in the dataset that were labeled with serious inaccuracies have been removed. However, due to the diverse nature of grasping methods for most objects, the overall FN error rate did not reach zero after ten iterations, primarily because we introduced new pseudo-generated labels. The accompanying figure clearly illustrates that with each subsequent iteration, the number of images erroneously categorized by the network decreases. This reduction in falsely labeled images can be attributed to the ongoing enhancement of the dataset. Initially, there were approximately 10,000 mislabeled images, eventually reduced to fewer than 4,000. These leftover 4,000 images represent objects with multiple interaction possibilities, highlighting the complexity of the NP-hard grasping problem. Overall, there is a diminishing trend in falsely labeled images as the number of iterations increases. Consequently, the time required for each iteration decreases as the dataset quality improves.
These findings collectively suggest a favorable advancement in the overall predictive performance of the network as the iterations progress.

\begin{figure}[htbp]
 \centering         
 \includegraphics[width=0.9\linewidth]{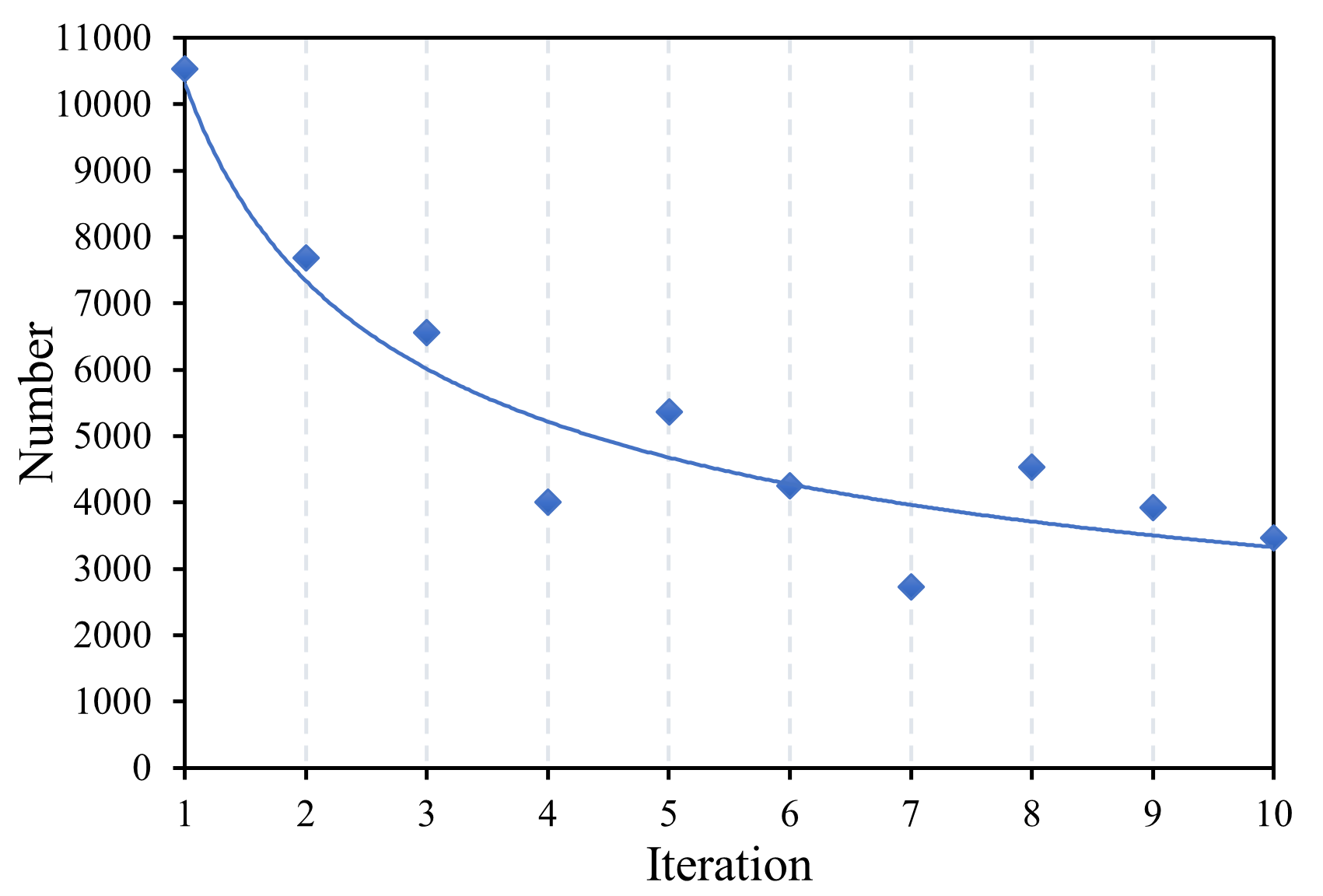}
 \caption{Iterative Process Flowchart for Quantity Variation Chart of False Images during Dataset Refinement.}
 \label{fig.p4}
 \end{figure}

 \begin{figure}[htbp]
 \centering         
 \includegraphics[width=0.9\linewidth]{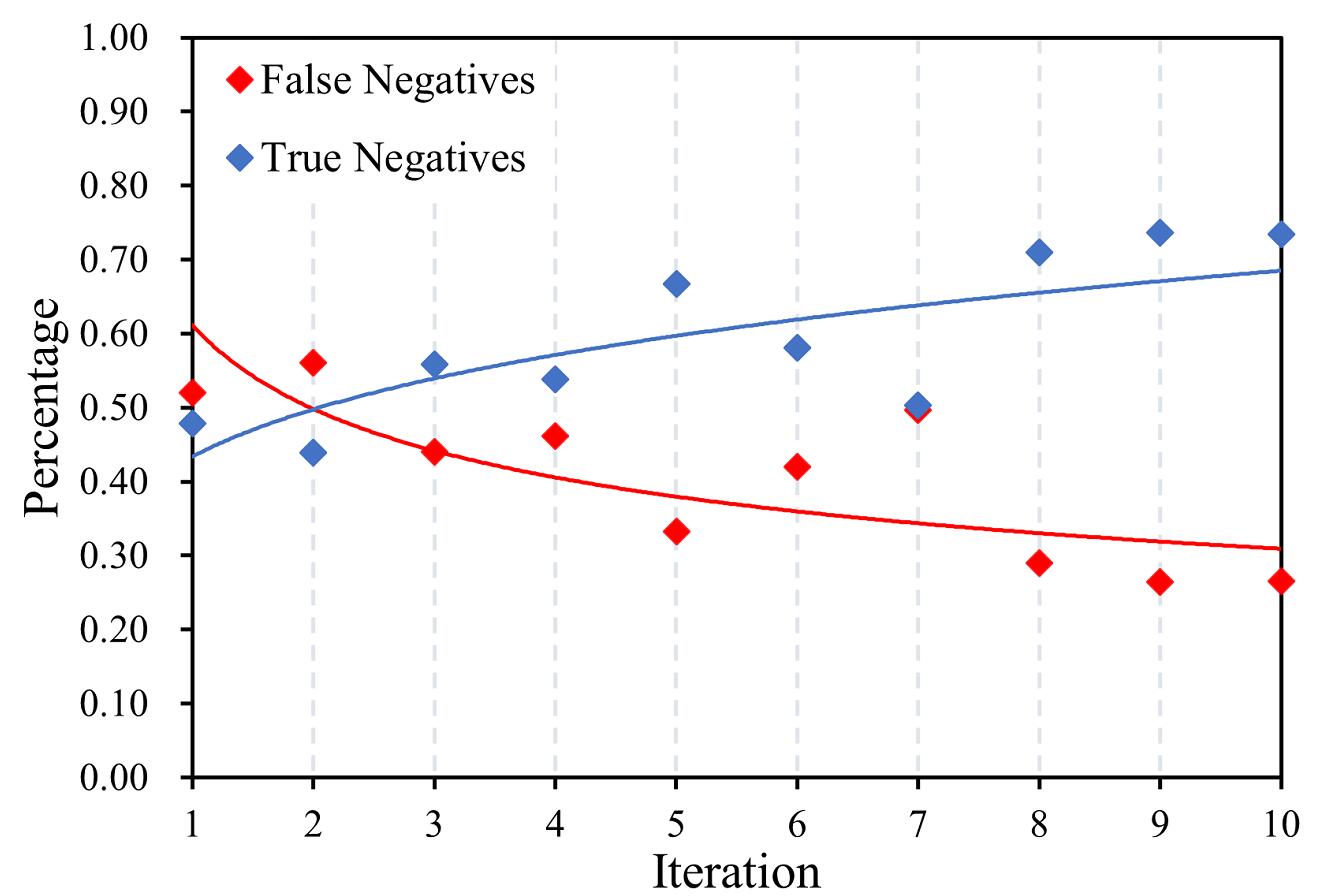}
 \caption{Iterative Process Flowchart for Proportion Chart of False Negative and True Negative during Dataset Refinement.}
 \label{fig.p5}
 \end{figure}
 
\subsection{Result and Analaysis} 
We provide a visual representation of the iterative analysis aimed at enhancing accuracy for various neural networks throughout the dataset refinement process. The figure illustrates each network performance on the test set using the dataset generated at each iteration, thereby validating the generalization capability of our method. Specifically, we trained and evaluated the following networks: MobileNet V2\cite{sandler2018mobilenetv2}, MobileNet V2\textsubscript{Pruning}, Squeeze\cite{SqueezeNet}, GG-CNN \cite{morrison2018closing}, GG-CNN2\cite{morrison2018closing}, ShuffleNet\cite{zhang2018shufflenet}, Xception\cite{chollet2017xception}, ResNet50\cite{he2016deep}, ResNet101\cite{he2016deep}, and ResNet152\cite{he2016deep}.

\begin{table}[htbp]
\centering
\caption{Iterative Process Data Graph for Accuracy Enhancement of Various Networks during Dataset Refinement (Unit: \%). The best result is highlighted in \textbf{bold},  while the second-best result is highlighted with an  \underline{underline}. }
\renewcommand{\arraystretch}{1.5}
\label{tb.1}
\begin{tabular}{lc|c|c|c|c}
\hline
\hline
\multicolumn{1}{c}{\multirow{2}{*}{\textbf{Network}}} & \multicolumn{5}{c}{\textbf{ Accuracy at each HIL iteration(\%)}} \\
\multicolumn{1}{c}{} & \textbf{0th} & \textbf{2nd} & \textbf{6th} & \textbf{8th} & \textbf{10th} \\
\hline
\textbf{ResNet50} & 84.27 & 86.6 & 89.46 & 91.03 & {91.37}\\
\hline
\textbf{ResNet101} & 84.59 & 86.19 & 87.44 & 89.63 & {91.35}\\
\hline
\textbf{ResNet152} & 84.26 & 85.01 & 87.17 & 88.73 & {90.3}\\
\hline
\textbf{Squeeze} & 91.97 & 92.92 & 93.35 & 94.75 & {94.76}\\

\hline
\textbf{ShuffleNet} & 91.24 & 91.78 & 92.93 & 93.6 & {94.85}\\
\hline
\textbf{Xception} & 89.03 & 92.52 & 93.00 & 94.36 & {95.27}\\

\hline
\textbf{GG-CNN} & 91.34 & 92.92 & 93.11 & 94.51 & {94.54}\\
\hline
\textbf{GG-CNN2} & 94.76 & 95.03 & 96.10 & 96.72 & \textbf{97.24}\\
\hline
\textbf{MobileNetV2} & 88.85 & 89.76 & 90.54 & 91.39 & {93.05} \\
\hline
\textbf{MobileNetV2\textsubscript{Pruning}} & 91.89 & 92.51 & 93.86 & 95.35 &  \underline{96.12}\\
\hline
\hline
\end{tabular}

\end{table}

Observing the Jacudar dataset from Table \ref{tb.1}, it becomes evident that more network layers do not necessarily yield better performance. For instance, ResNet152 achieves only 90\% accuracy, while its less complex variant, ResNet50, attains 91.37\% accuracy. This suggests that the standard ResNet152 architecture proves excessively deep for this dataset of 50,000 images, resulting in a substantial degree of overfitting
. Most other datasets achieve approximately 95 percent accuracy as a benchmark, while GG-CNN2, optimized for this smaller dataset, achieves a remarkable 97 percent accuracy.

To assess the potential of slight overfitting, we pruned the MobileNetV2 backbone by removing redundant convolution layers, resulting in a variant method named MobileNetV2\textsubscript{Pruning}. The newer punning version resulted in a three \% performance improvement compared to the original MobileNetV2. 
In summary, while this dataset demonstrates limited compatibility with deeper networks, our proposed solution consistently enhances accuracy through improved annotation and label learning. However, when dealing with datasets featuring complex backgrounds, we believe MobileNetV2 can outperform the GG-CNN series of models. However, that will be left for some further work to verify.

\begin{figure}[htbp]
 \centering         
 \includegraphics[width=0.9\linewidth]{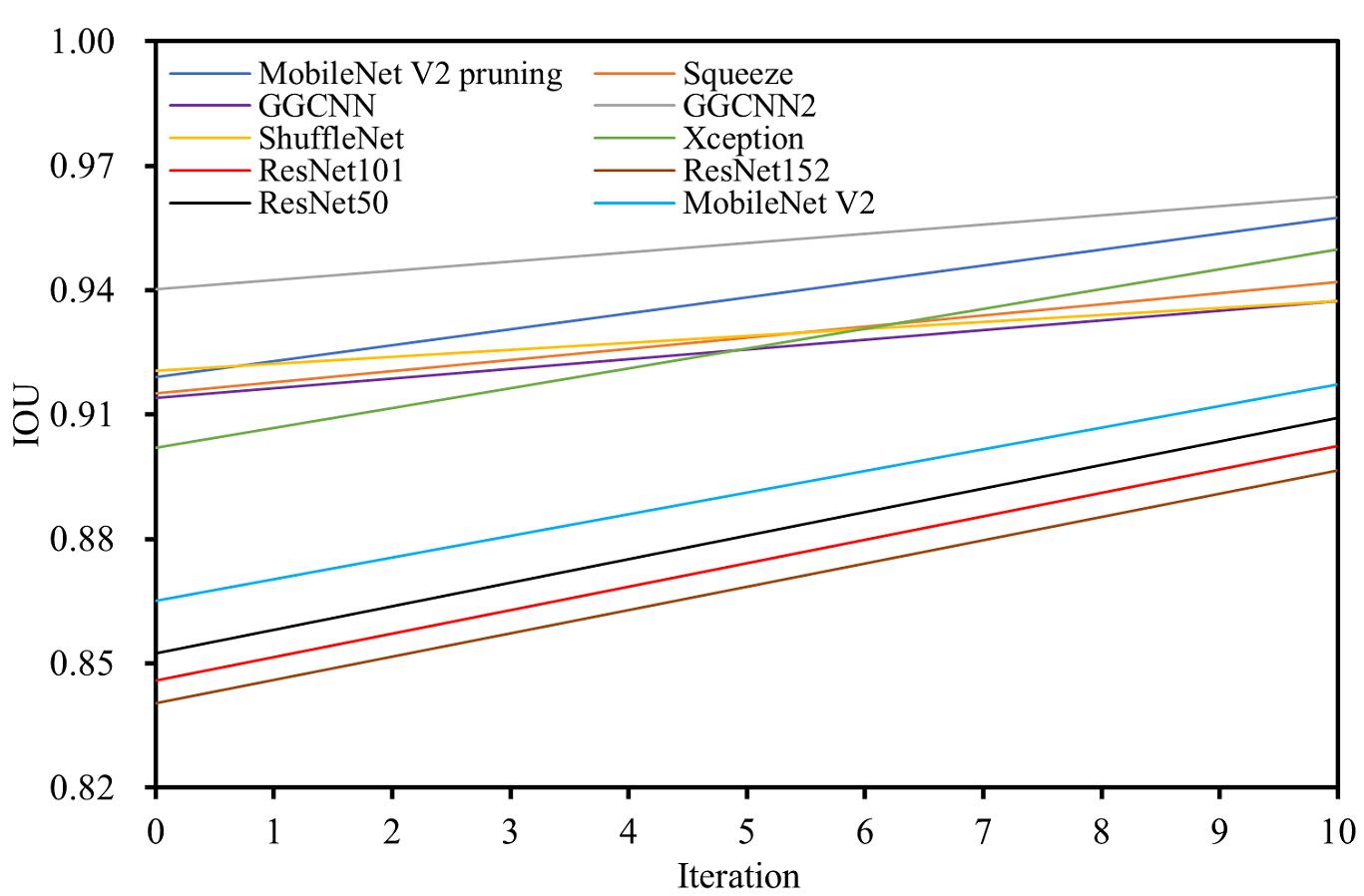}
 \caption{Iterative Process Flowchart for Accuracy Enhancement of Various Networks during Dataset Refinement.}
 \label{fig.p6}
 \end{figure}

In Fig. \ref{fig.p6}, trend lines depict that, without modifying the network structures, all networks demonstrate improved accuracy compared to their performance before the iterations. However, it's essential to acknowledge that network training accuracy can be influenced by factors like computer hardware, learning rate adjustment strategies, etc. As a result, there may be cases where the scores after specific iterations are lower than before.

Due to constraints in our GPU resources, conducting multiple experiments and averaging the results for each network and iteration was not feasible. Instead, we present the outcomes of a single experiment along with the fitted curves. We believe that achieving more robust visualization results could be attainable by incorporating multiple experiments and averaging the outcomes.

 \begin{figure}[htbp]
 \centering         
 \includegraphics[width=0.9\linewidth]{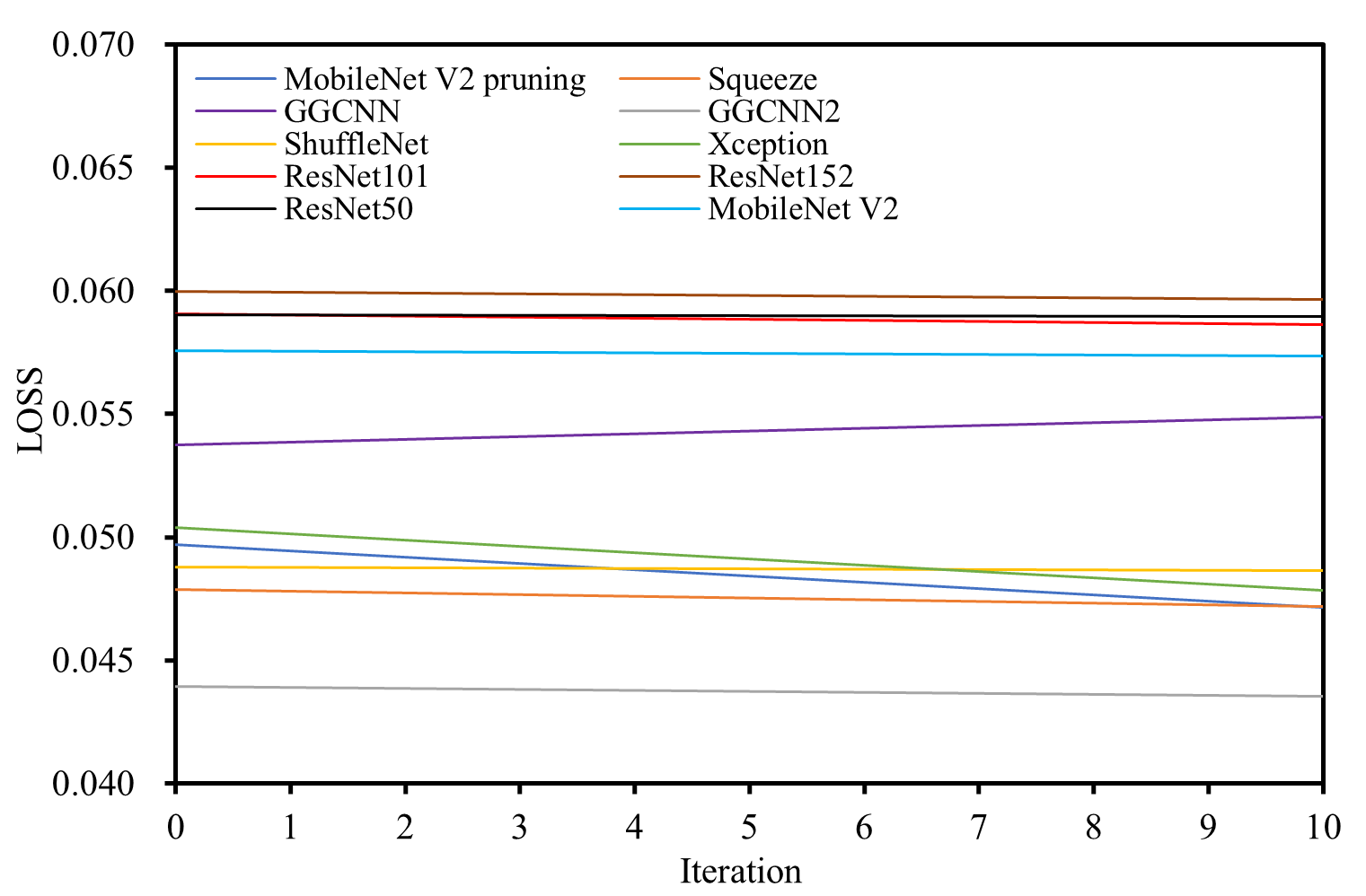}
 \caption{Iterative Process Flowchart for Loss Reduction of Various Networks during Dataset Refinement.}
 \label{fig.p7}
 \end{figure}

This iterative analysis focuses on loss reduction for various networks during the dataset refinement process. We employ neural networks similar to those depicted in Fig. \ref{fig.p7}  and track the loss on the training set. The results reveal a common trend among most networks, demonstrating a gradual decrease in loss as the number of iterations increases. However, the GG-CNN network, characterized by its simple and less mature network structure, remains in an underfitting state. This hinders its ability to capture updated knowledge, resulting in a lack of a discernible downward trend in the loss curve.

\begin{figure}[htbp]
    \centering         
    \includegraphics[width=\linewidth]{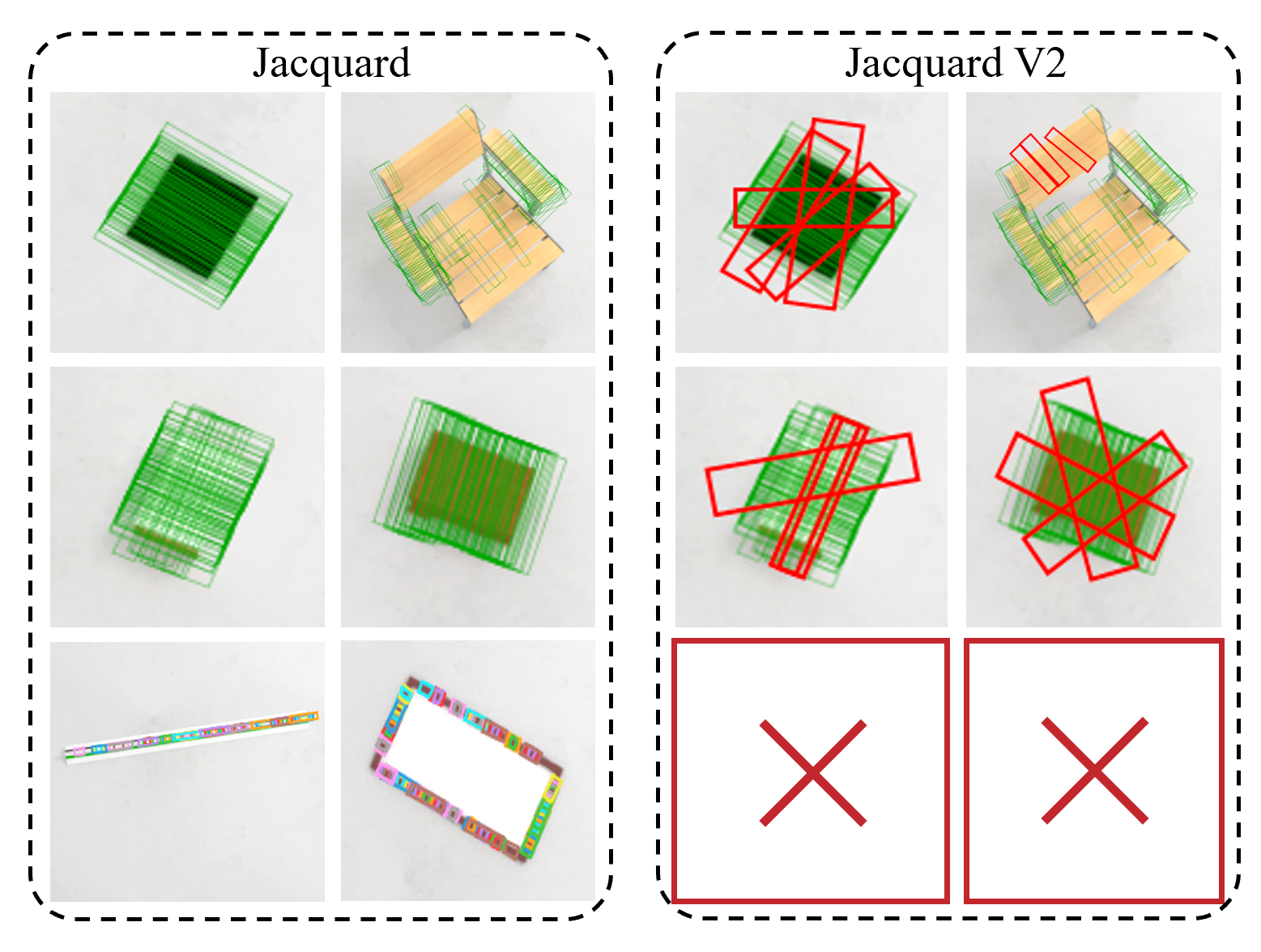}
    \caption{In the left image set, annotations from the original dataset are displayed. In the right image set, after ten rounds of improvement, it becomes evident that many possible grasping directions have been added, aligning them more closely with the human nature approach to grasping.
}
    \label{fig.p8}
\end{figure}
In summary, after ten iterations, we have observed substantial enhancements in label quality, with the addition of new labels and the correction of serious mistakes, as illustrated in Figure \ref{fig.p8}. This signifies our effective efforts in creating an improved dataset for the entire research community.

\section{Known issue and limitations}
While our approach shows promise, it is important to acknowledge its limitations. During the HIL reviewing phase, we had to remove poorly annotated image portions, resulting in irreversible information loss. Additionally, our system's performance evaluation is based on a single dataset, and issues may vary with newer datasets. Furthermore, the vast diversity of grasping methods available makes it challenging to eliminate FN entirely. Our method reduces errors but does not entirely eliminate them. One potential avenue for improvement could involve using large language models to incorporate prior knowledge, though this falls outside the scope of our current work.
\section{conclusion}
This paper presents a novel HIL approach to creating the Jacquard V2 dataset, employing a meticulous selection and annotation process that enhances data quality and consistency. We conducted manual examinations to verify the accuracy of negative predictions and removed 2,884 images that violated grasping principles, resulting in a refined dataset with precise ground truth labels totaling 51,601. This dataset, trained with the same network architecture, led to a remarkable 7.1\% increase in test set accuracy compared to the original Jacquard Grasp Dataset\cite{depierre2018jacquard}. Our method not only rectifies the shortcomings of incomplete annotations through the HIL approach but also sheds light on the balance between data volume and model complexity in grasping tasks. These contributions significantly advance the field of robotic visual grasping and enhance machine perception and manipulation capabilities. To facilitate further progress, we are committed to open-sourcing both the benchmark and the refined dataset for the research community.



\bibliographystyle{IEEEtran}
\bibliography{mybib}

\begin{thebibliography}{10}
\providecommand{\url}[1]{#1}
\csname url@samestyle\endcsname
\providecommand{\newblock}{\relax}
\providecommand{\bibinfo}[2]{#2}
\providecommand{\BIBentrySTDinterwordspacing}{\spaceskip=0pt\relax}
\providecommand{\BIBentryALTinterwordstretchfactor}{4}
\providecommand{\BIBentryALTinterwordspacing}{\spaceskip=\fontdimen2\font plus
\BIBentryALTinterwordstretchfactor\fontdimen3\font minus
  \fontdimen4\font\relax}
\providecommand{\BIBforeignlanguage}[2]{{%
\expandafter\ifx\csname l@#1\endcsname\relax
\typeout{** WARNING: IEEEtran.bst: No hyphenation pattern has been}%
\typeout{** loaded for the language `#1'. Using the pattern for}%
\typeout{** the default language instead.}%
\else
\language=\csname l@#1\endcsname
\fi
#2}}
\providecommand{\BIBdecl}{\relax}
\BIBdecl

\bibitem{depierre2018jacquard}
A.~Depierre, E.~Dellandr{\'e}a, and L.~Chen, ``Jacquard: A large scale dataset
  for robotic grasp detection,'' in \emph{2018 IEEE/RSJ International
  Conference on Intelligent Robots and Systems (IROS)}.\hskip 1em plus 0.5em
  minus 0.4em\relax IEEE, 2018, pp. 3511--3516.

\bibitem{mahler2017dex}
J.~Mahler, J.~Liang, S.~Niyaz, M.~Laskey, R.~Doan, X.~Liu, J.~A. Ojea, and
  K.~Goldberg, ``Dex-net 2.0: Deep learning to plan robust grasps with
  synthetic point clouds and analytic grasp metrics,'' \emph{arXiv preprint
  arXiv:1703.09312}, 2017.

\bibitem{xiang2017posecnn}
Y.~Xiang, T.~Schmidt, V.~Narayanan, and D.~Fox, ``Posecnn: A convolutional
  neural network for 6d object pose estimation in cluttered scenes,''
  \emph{arXiv preprint arXiv:1711.00199}, 2017.

\bibitem{rusu2010semantic}
R.~B. Rusu, ``Semantic 3d object maps for everyday manipulation in human living
  environments,'' \emph{KI-K{\"u}nstliche Intelligenz}, vol.~24, pp. 345--348,
  2010.

\bibitem{tang2012textured}
J.~Tang, S.~Miller, A.~Singh, and P.~Abbeel, ``A textured object recognition
  pipeline for color and depth image data,'' in \emph{2012 IEEE International
  Conference on Robotics and Automation}.\hskip 1em plus 0.5em minus
  0.4em\relax IEEE, 2012, pp. 3467--3474.

\bibitem{zaheer2020big}
M.~Zaheer, G.~Guruganesh, K.~A. Dubey, J.~Ainslie, C.~Alberti, S.~Ontanon,
  P.~Pham, A.~Ravula, Q.~Wang, L.~Yang \emph{et~al.}, ``Big bird: Transformers
  for longer sequences,'' \emph{Advances in neural information processing
  systems}, vol.~33, pp. 17\,283--17\,297, 2020.

\bibitem{tenorth2009tum}
M.~Tenorth, J.~Bandouch, and M.~Beetz, ``The tum kitchen data set of everyday
  manipulation activities for motion tracking and action recognition,'' in
  \emph{2009 IEEE 12th International Conference on Computer Vision Workshops,
  ICCV Workshops}.\hskip 1em plus 0.5em minus 0.4em\relax IEEE, 2009, pp.
  1089--1096.

\bibitem{gao2014jhu}
Y.~Gao, S.~S. Vedula, C.~E. Reiley, N.~Ahmidi, B.~Varadarajan, H.~C. Lin,
  L.~Tao, L.~Zappella, B.~B{\'e}jar, D.~D. Yuh \emph{et~al.}, ``Jhu-isi gesture
  and skill assessment working set (jigsaws): A surgical activity dataset for
  human motion modeling,'' in \emph{MICCAI workshop: M2cai}, vol.~3, no.~3,
  2014.

\bibitem{castrejon2017annotating}
L.~Castrejon, K.~Kundu, R.~Urtasun, and S.~Fidler, ``Annotating object
  instances with a polygon-rnn,'' in \emph{Proceedings of the IEEE conference
  on computer vision and pattern recognition}, 2017, pp. 5230--5238.

\bibitem{nguyen2022ntu}
T.-M. Nguyen, S.~Yuan, M.~Cao, Y.~Lyu, T.~H. Nguyen, and L.~Xie, ``Ntu viral: A
  visual-inertial-ranging-lidar dataset, from an aerial vehicle viewpoint,''
  \emph{The International Journal of Robotics Research}, vol.~41, no.~3, pp.
  270--280, 2022.

\bibitem{bruls2018mark}
T.~Bruls, W.~Maddern, A.~A. Morye, and P.~Newman, ``Mark yourself: Road marking
  segmentation via weakly-supervised annotations from multimodal data,'' in
  \emph{2018 IEEE International Conference on Robotics and Automation
  (ICRA)}.\hskip 1em plus 0.5em minus 0.4em\relax IEEE, 2018, pp. 1863--1870.

\bibitem{marion2018label}
P.~Marion, P.~R. Florence, L.~Manuelli, and R.~Tedrake, ``Label fusion: A
  pipeline for generating ground truth labels for real rgbd data of cluttered
  scenes,'' in \emph{2018 IEEE International Conference on Robotics and
  Automation (ICRA)}.\hskip 1em plus 0.5em minus 0.4em\relax IEEE, 2018, pp.
  3235--3242.

\bibitem{barnes2017find}
D.~Barnes, W.~Maddern, and I.~Posner, ``Find your own way: Weakly-supervised
  segmentation of path proposals for urban autonomy,'' in \emph{2017 IEEE
  International Conference on Robotics and Automation (ICRA)}.\hskip 1em plus
  0.5em minus 0.4em\relax IEEE, 2017, pp. 203--210.

\bibitem{bousmalis2018using}
K.~Bousmalis, A.~Irpan, P.~Wohlhart, Y.~Bai, M.~Kelcey, M.~Kalakrishnan,
  L.~Downs, J.~Ibarz, P.~Pastor, K.~Konolige \emph{et~al.}, ``Using simulation
  and domain adaptation to improve efficiency of deep robotic grasping,'' in
  \emph{2018 IEEE international conference on robotics and automation
  (ICRA)}.\hskip 1em plus 0.5em minus 0.4em\relax IEEE, 2018, pp. 4243--4250.

\bibitem{eppner2021acronym}
C.~Eppner, A.~Mousavian, and D.~Fox, ``Acronym: A large-scale grasp dataset
  based on simulation,'' in \emph{2021 IEEE International Conference on
  Robotics and Automation (ICRA)}.\hskip 1em plus 0.5em minus 0.4em\relax IEEE,
  2021, pp. 6222--6227.

\bibitem{kappler2015leveraging}
D.~Kappler, J.~Bohg, and S.~Schaal, ``Leveraging big data for grasp planning,''
  in \emph{2015 IEEE international conference on robotics and automation
  (ICRA)}.\hskip 1em plus 0.5em minus 0.4em\relax IEEE, 2015, pp. 4304--4311.

\bibitem{suchi2019easylabel}
M.~Suchi, T.~Patten, D.~Fischinger, and M.~Vincze, ``Easylabel: A
  semi-automatic pixel-wise object annotation tool for creating robotic rgb-d
  datasets,'' in \emph{2019 International Conference on Robotics and Automation
  (ICRA)}.\hskip 1em plus 0.5em minus 0.4em\relax IEEE, 2019, pp. 6678--6684.

\bibitem{handa2016scenenet}
A.~Handa, V.~P{\u{a}}tr{\u{a}}ucean, S.~Stent, and R.~Cipolla, ``Scenenet: An
  annotated model generator for indoor scene understanding,'' in \emph{2016
  IEEE International Conference on Robotics and Automation (ICRA)}.\hskip 1em
  plus 0.5em minus 0.4em\relax IEEE, 2016, pp. 5737--5743.

\bibitem{krizhevsky2012imagenet}
A.~Krizhevsky, I.~Sutskever, and G.~E. Hinton, ``Imagenet classification with
  deep convolutional neural networks,'' \emph{Advances in neural information
  processing systems}, vol.~25, 2012.

\bibitem{ioffe2015batch}
S.~Ioffe and C.~Szegedy, ``Batch normalization: Accelerating deep network
  training by reducing internal covariate shift,'' in \emph{International
  conference on machine learning}.\hskip 1em plus 0.5em minus 0.4em\relax pmlr,
  2015, pp. 448--456.

\bibitem{ioffe2017batch}
S.~Ioffe, ``Batch renormalization: Towards reducing minibatch dependence in
  batch-normalized models,'' \emph{Advances in neural information processing
  systems}, vol.~30, 2017.

\bibitem{ba2016layer}
J.~L. Ba, J.~R. Kiros, and G.~E. Hinton, ``Layer normalization,'' \emph{arXiv
  preprint arXiv:1607.06450}, 2016.

\bibitem{tan2023human}
T.~Tan, ``A human-in-the-loop robot grasping system with grasp quality
  refinement,'' Ph.D. dissertation, University of South Florida, 2023.

\bibitem{morrison2018closing}
D.~Morrison, P.~Corke, and J.~Leitner, ``{Closing the Loop for Robotic
  Grasping: A Real-time, Generative Grasp Synthesis Approach},'' in
  \emph{Proc.\ of Robotics: Science and Systems (RSS)}, 2018.

\bibitem{hara2017designing}
K.~Hara, R.~Vemulapalli, and R.~Chellappa, ``Designing deep convolutional
  neural networks for continuous object orientation estimation,'' \emph{arXiv
  preprint arXiv:1702.01499}, 2017.

\bibitem{sandler2018mobilenetv2}
M.~Sandler, A.~Howard, M.~Zhu, A.~Zhmoginov, and L.-C. Chen, ``Mobilenetv2:
  Inverted residuals and linear bottlenecks,'' in \emph{Proceedings of the IEEE
  conference on computer vision and pattern recognition}, 2018, pp. 4510--4520.

\bibitem{SqueezeNet}
F.~N. Iandola, S.~Han, M.~W. Moskewicz, K.~Ashraf, W.~J. Dally, and K.~Keutzer,
  ``Squeezenet: Alexnet-level accuracy with 50x fewer parameters and $<$0.5mb
  model size,'' \emph{arXiv:1602.07360}, 2016.

\bibitem{zhang2018shufflenet}
X.~Zhang, X.~Zhou, M.~Lin, and J.~Sun, ``Shufflenet: An extremely efficient
  convolutional neural network for mobile devices,'' in \emph{Proceedings of
  the IEEE conference on computer vision and pattern recognition}, 2018, pp.
  6848--6856.

\bibitem{chollet2017xception}
F.~Chollet, ``Xception: Deep learning with depthwise separable convolutions,''
  in \emph{Proceedings of the IEEE conference on computer vision and pattern
  recognition}, 2017, pp. 1251--1258.

\bibitem{he2016deep}
K.~He, X.~Zhang, S.~Ren, and J.~Sun, ``Deep residual learning for image
  recognition,'' in \emph{Proceedings of the IEEE conference on computer vision
  and pattern recognition}, 2016, pp. 770--778.

\end{thebibliography}

\end{document}